Original Paper

# Predicting Intensive Care Unit Length of Stay and Mortality Using Patient Vital Signs: Machine Learning Model Development and Validation

Khalid Alghatani[1], PhD; Nariman Ammar[2], PhD; Abdelmounaam Rezgui[3], PhD; Arash Shaban-Nejad[2], PhD

[1]Department of Computer Science and Engineering, New Mexico Institute of Mining and Technology, Socorro, NM, United States
[2]Oak Ridge National Laboratory Center for Biomedical Informatics, Department of Pediatrics, College of Medicine, The University of Tennessee Health Science Center, Memphis, TN, United States
[3]School of Information Technology, Illinois State University, Normal, IL, United States

**Corresponding Author:**
Khalid Alghatani, PhD
Department of Computer Science and Engineering
New Mexico Institute of Mining and Technology
801 Leroy Pl
Socorro, NM, 87801
United States
Phone: 1 5057204644
Email: khalid.alghatani@student.nmt.edu

## Abstract

**Background:** Patient monitoring is vital in all stages of care. In particular, intensive care unit (ICU) patient monitoring has the potential to reduce complications and morbidity, and to increase the quality of care by enabling hospitals to deliver higher-quality, cost-effective patient care, and improve the quality of medical services in the ICU.

**Objective:** We here report the development and validation of ICU length of stay and mortality prediction models. The models will be used in an intelligent ICU patient monitoring module of an Intelligent Remote Patient Monitoring (IRPM) framework that monitors the health status of patients, and generates timely alerts, maneuver guidance, or reports when adverse medical conditions are predicted.

**Methods:** We utilized the publicly available Medical Information Mart for Intensive Care (MIMIC) database to extract ICU stay data for adult patients to build two prediction models: one for mortality prediction and another for ICU length of stay. For the mortality model, we applied six commonly used machine learning (ML) binary classification algorithms for predicting the discharge status (survived or not). For the length of stay model, we applied the same six ML algorithms for binary classification using the median patient population ICU stay of 2.64 days. For the regression-based classification, we used two ML algorithms for predicting the number of days. We built two variations of each prediction model: one using 12 baseline demographic and vital sign features, and the other based on our proposed quantiles approach, in which we use 21 extra features engineered from the baseline vital sign features, including their modified means, standard deviations, and quantile percentages.

**Results:** We could perform predictive modeling with minimal features while maintaining reasonable performance using the quantiles approach. The best accuracy achieved in the mortality model was approximately 89% using the random forest algorithm. The highest accuracy achieved in the length of stay model, based on the population median ICU stay (2.64 days), was approximately 65% using the random forest algorithm.

**Conclusions:** The novelty in our approach is that we built models to predict ICU length of stay and mortality with reasonable accuracy based on a combination of ML and the quantiles approach that utilizes only vital signs available from the patient's profile without the need to use any external features. This approach is based on feature engineering of the vital signs by including their modified means, standard deviations, and quantile percentages of the original features, which provided a richer dataset to achieve better predictive power in our models.

XSL•FO
RenderX

## Introduction

### Background

Precision observation and assessment are crucial tasks for "achieving an early diagnosis, informed planning, reflecting on the suitability of treatment options, information exchanging, and designing better health interventions" [1]. The use of artificial intelligence–based solutions to improve health care services is increasing [2] and patient monitoring is now an integral part of clinical intelligence [3]. The intensive care unit (ICU) is one of the most critical and resource-intensive units in hospitals, and ICU patient monitoring and continuous clinical surveillance have the potential to reduce morbidity and improve the quality of care. Therefore, hospitals often seek solutions that enable reducing waste and wait times, while increasing service efficiencies, accuracy, and productivity [2]. One of the issues in current monitoring approaches is that the data are collected via sensing devices and sent to remote diagnostic testing facilities for further, often manual or semiautomated, interpretation by a health care professional. Thus, there is a need for intelligent solutions for ICU patient monitoring that require minimal human intervention and that can monitor the health status of patients, and generate timely alerts, maneuver guidance, and reports whenever adverse medical conditions are anticipated.

In our previous work [4], we proposed an Intelligent Remote Patient Monitoring (IRPM) framework (Figure 1) that consists of three modules: (i) an out-of-hospital module that utilizes data collected via wearable devices (eg, Apple Watch and SleepO2); (ii) a decision support module that generates reports; and (iii) an intelligent ICU patient monitoring module, which utilizes data collected from ICUs. We here focus on the latter module.

**Figure 1.** Intelligent Remote Patient Monitoring (IRPM) framework. IICUPM: intelligent intensive care unit patient monitoring.

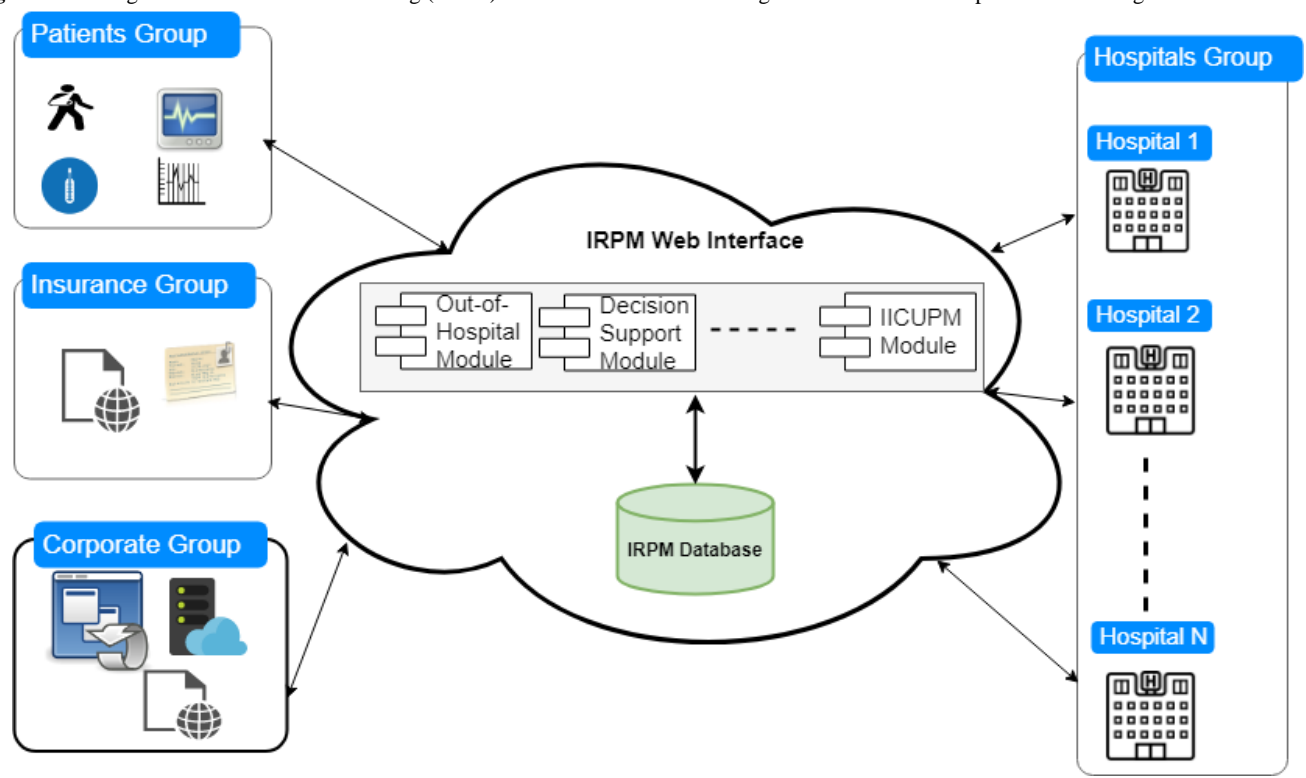

The IRPM framework is intended to serve as a global web service interface that exposes the different framework functionalities to hospitals, hospital managers, insurance companies, and other decision makers, including the host organizations that operate and maintain the IRPM system. The intelligent ICU patient monitoring functionality of the service performs analytics of the data exchanged between ICUs and the core IRPM system, and provides the different stakeholders with the analysis results in the form of timely and early warnings.

Three main factors impact the quality of prediction models: (1) the target patient population [5], (2) methods used for data fusion, and (3) algorithm type. Different populations lead to different prediction results. Moreover, different ways of combining information from physiological variables lead to various outcome measures. The IRPM framework is intended to be hosted in the cloud since the intelligent ICU patient monitoring module aims at applying machine learning (ML) within an architecture that allows any user (regardless of whether or not they are sick) as well as any hospital system to use the framework. Since most of the used physiological variables are often obtained inside and outside hospitals, the framework will enable performing continuous patient monitoring. Therefore, we built ML models by utilizing features that are easy to obtain outside the hospital setting, and we avoided features that are sophisticated and require high-level medical equipment.

### Related Works

There has been some research effort toward developing ML models for predicting ICU-related outcomes [6-8]. McCarthy et al [9] performed a study on ICU mortality prediction in which they compared sliding-window predictors with recurrent predictors to classify patient state of health from ICU multivariate time-series data. They reported slightly improved performance for the recurrent neural network. Moreover, Zhu et al [10] proposed an ICU mortality prediction algorithm combining the bidirectional long short-term memory (LSTM) model with supervised learning. They trained and evaluated the LSTM model using 4000 ICU patients. They also performed a comparative analysis, which identified that their proposed method significantly outperformed several baseline methods.

A few studies have also focused on developing and validating ML models for predicting ICU-related outcomes using the Medical Information Mart for Intensive Care (MIMIC) database. Most of these works have used an exhaustive list of features to achieve higher accuracy in their models. Johnson and colleagues [11-13] developed models for predicting ICU mortality, achieving an area under the receiver operating characteristic (ROC) curve (AUROC) of 0.92 using a total of 148 features [12] and an AUROC of 0.86 using a range of features, including standard statistical descriptors [13]. Lehman et al [14] used basic physiological variables and applied the Simplified Acute Physiology Score (SAPS-I) algorithm to predict mortality, which achieved an AUROC of 0.72. Using the Cohen standardized mean and coefficient, Tyler et al [15] assessed the differences between ICU lab values, which were used to predict ICU length of stay (LOS) and mortality. Harutyunyan et al [16] selected 17 clinical variables to build a binary LOS model to predict whether a patient will stay in the ICU for a long (≥7 days) or short (<7 days) period with 84% accuracy. Gentimis et al [17] used several inputs from seven tables to build an LOS model to predict whether a patient will stay in the ICU for a long (>5 days) or short (≤5 days) period using neural networks, with around 80% accuracy; they removed patients who stayed in the ICU longer than 20 days. Bertsimas et al [18] used several static and dynamic variables (eg, general admission data, lab results, medical orders, pharmacy data, diagnosis codes, and notes) and different classification methods to predict different LOS with accuracy in the >80% range.

Some works have focused on developing ML models to be used in clinical information systems that assist in ICU discharge planning. Badawi and Breslow [19] developed and validated two models for predicting risks of death and readmission within 48 hours of ICU discharge. They used eICU Research Institute data from more than 400 ICUs and performed multivariate logistic regression (MLR) with 59 different features, including patient demographics, ICU admission diagnosis, admission severity of illness, intensive care interventions, complications occurring during the ICU stay, lab values, and physiological variables recorded within the last 24 hours of the ICU stay. They calibrated their models across deciles of risk, and their mortality model accurately discriminated between patients who would and would not experience a complication as early as 4 days before ICU discharge. However, to the best of our knowledge, predicting the LOS based on the population's median ICU patient stay using only vital signs and demographic attributes from MIMIC data has not been studied to date.

### Objective

We here propose a new approach that focuses on the most critical observations in a patient's profile. The novelty of the approach lies in its ability to predict outcomes with reasonable accuracy by utilizing only vital signs that exist in the patient's profile without having prior knowledge about a patient's medical conditions or diagnoses. The approach enriches the original vital sign measures by adding extra features pertaining to their modified means, modified SDs, and quantile percentages. We evaluated the proposed approach (ie, the quantiles approach) in comparison to a baseline approach that uses the entire range of observations. We then applied both approaches to develop and validate two prediction models: (i) one focusing on classifying ICU mortality rate (survival or no survival), and (ii) another focusing on predicting the LOS in the ICU using public data from the MIMIC database.

## Methods

### Study Population and Data Extraction

We used MIMIC-III (v1.4) [7], a publicly available ICU adult patient database that spans 11 years between 2001 and 2012. MIMIC-III has data for 53,423 distinct hospital admissions, including nearly 500 million rows in 26 tables. The database comprises features, including patient demographics, laboratory test results, medical reports, and results from imaging studies. To meet Health Insurance Portability and Accountability Act requirements, approximate ages for patients who are more than 89 years are reported by shifting their date of birth.

Figure 2 illustrates the data extraction pipeline of ICU stays data from the MIMIC database. We started with 61,532 total ICU stay encounters. In each hospital admission, a patient could have stayed in the ICU more than once. We performed this study based on unique ICU stays rather than unique patient identifiers since our goal was to predict mortality and LOS without having prior knowledge about patients' medical conditions or diagnoses.

For patients who stayed in the ICU for at least 1 day, we considered their data for only the first day. The population's median ICU LOS was 2.64 days, and therefore we discarded data from patients who stayed in the ICU for less than 1 day, which resulted in a total of 45,254 unique ICU stays. For each ICU stay, we ran separate SQL queries to extract the patients' vital sign measurements, and height and weight features from the total 61,532 encounters. We focused on six vital sign features (body temperature, heart rate, respiration rate, systolic blood pressure, diastolic blood pressure, and oxygen saturation [$SpO_2$]) along with glucose level. The total number of ICU stays for which vital sign features were available was 59,241. We extracted four demographic features (weight, height, age, and gender). We then performed consecutive inner joins between the results of the three queries; thus, the total ICU stays reduced to 44,626 unique ICU stays.

**Figure 2.** Data extraction pipeline from the Medical Information Mart for Intensive Care (MIMIC) database. ICU: intensive care unit.

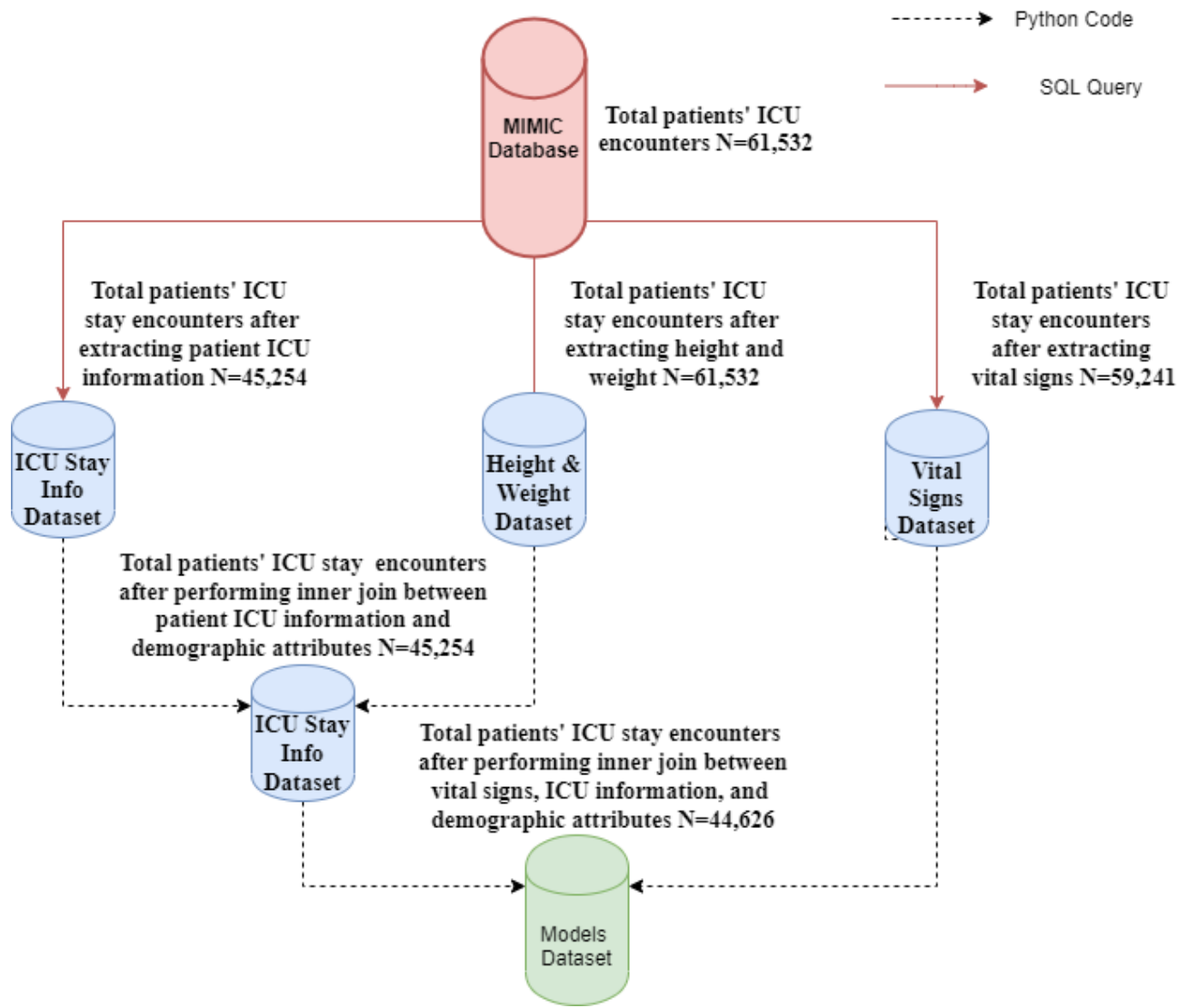

### Data Preprocessing

To enhance the accuracy of the predictive models, we eliminated extreme, trivial, and negative observations within each vital sign feature. The percentage of missing data was relatively low (less than 1% for heart rate, respiration rate, systolic blood pressure, diastolic blood pressure, $SpO_2$, and glucose level, and less than 2% for body temperature). Given the low percentage of missing values and the fact that vital signs are numerical values that are typically normally distributed [20], we filled missing values of vital sign observations using the mean.

### Model and Variable Selection

We built two main prediction models: in-hospital mortality and LOS for each ICU admission. Table 1 defines the outcome variables in both models. The outcome variable for the mortality model was in-hospital mortality, which reduces to a binary classification problem with two classes: predicting a patient to survive or not. The dataset has a classification imbalance problem since the in-hospital mortality percentage was 11.897%, whereas the patient survival percentage was 88.103%. The outcome for the LOS model was the number of days a patient stayed in the ICU. Half of the population spent 2.64 days in the ICU, which led us to follow two approaches for classification. In the first approach, we followed a binary classification strategy by defining two classes with an equal number of observations by considering 2.64 as a threshold. The first class predicts that a patient will stay in the ICU for 2.64 days or less, and the second class predicts that a patient will stay in the ICU for more than 2.64 days. In the second approach, we followed a regression-based classification strategy by considering the predicted outcome as a continuous variable.

We built two variations of each model: one using the baseline approach and another using the proposed quantiles approach. The models built with the baseline approach used the six vital sign features, glucose, and the five demographic features as predictor variables (Table 2). The models built with the quantiles approach used the same 12 baseline predictor variables, and augmented them with extra modified features. We discuss each model variation separately below.

**Table 1.** Descriptive statistics for outcome variables in the two models.

| Model | Operationalization | Values |
| --- | --- | --- |
| In-hospital mortality (binary classification) | 0: survival; 1: nonsurvival | 0: 11.897%; 1: 88.10 3% |
| **Length of stay (LOS)** | | |
| Binary classification | 0: LOS≤2.636 days; 1: LOS>2.636 days | 0: 50%; 1: 50% |
| Regression-based classification | Number of days in intensive care unit | Mean 4.74959 (SD 6.49982) |

**Table 2.** Descriptive statistics for baseline model predictors (N=44,626).

| Input variables | Measurement | Value |
| --- | --- | --- |
| HeartRate_mean | Heart rate (beats/minute), mean (SD) | 85.99 (15.59) |
| sysbp_mean | Arterial systolic blood pressure (mmHg) mean (SD) | 118.75 (16.90) |
| diasbp_mean | Arterial diastolic blood pressure (mmHg), mean (SD) | 60.47 (10.89) |
| RespRate_mean | Respiratory rate (breaths/minute), mean (SD) | 18.93 (4.05) |
| Tempc_mean | Body temperature (°C), mean (SD) | 36.84 (0.62) |
| $Spo_2$_mean | Peripheral oxygen saturation (%), mean | 97.27 |
| Glucose_mean | Blood glucose (mg/dL), mean (SD) | 138.74 (41.86) |
| Age | Age (years), mean (SD) | 64.35 (16.87) |
| GenderM | Male population, n (%) | 25,241 (56.56) |
| GenderF | Female population, n (%) | 19,385 (43.44) |
| Height | Patient height (cm), mean (SD) | 160.66 (11.76) |
| Weight | Patient weight (kg), mean (SD) | 80.45 (23.47) |

## Baseline Approach

Table 2 shows the descriptive statistics for the predictor variables used in the baseline approach: the patients' vital signs for the first day and the demographic variables. The population had a slight majority of men with a mean age of 64.35 years.

Pearson correlation coefficients among the vital sign variables in the baseline approach (Table 3) showed weak correlations between the variables, except between systolic and diastolic blood pressure.

**Table 3.** Pearson correlation coefficients among vital signs of the baseline model.

| Variable | Heart rate | Systolic BP[a] | Diastolic BP | Respiration rate | Body temperature | $SpO_2$[b] | Glucose |
| --- | --- | --- | --- | --- | --- | --- | --- |
| Heart rate | 1 | –0.104 | 0.211 | 0.326 | 0.268 | –0.099 | 0.063 |
| Systolic BP | –0.104 | 1 | 0.524 | –0.032 | 0.065 | 0.045 | 0.063 |
| Diastolic BP | 0.211 | 0.524 | 1 | 0.0257 | 0.065 | –0.0148 | 0.0142 |
| Respiration rate | 0.326 | –0.032 | 0.0257 | 1 | 0.118 | –0.259 | 0.069 |
| Body temperature | 0.268 | 0.065 | 0.0335 | 0.118 | 1 | 0.051 | –0.022 |
| $SPO_2$ | –0.099 | 0.045 | –0.0148 | –0.259 | 0.051 | 1 | –0.048 |
| Glucose | 0.063 | 0.078 | 0.0142 | 0.069 | –0.022 | –0.048 | 1 |

[a]BP: blood pressure.
[b]$SpO_2$: oxygen saturation.

## Quantiles Approach

When dealing with sequential data, observations that are far from the median are often ignored. We argue that a patient's deteriorating condition often comes with a high or low level of measurement. Thus, these observations are essential as they report the point at which the patient's health status changes dramatically. We propose the notion of the "quantiles approach," in which we perform feature engineering by emphasizing the high and low quantiles of a patient sample. Figure 3 demonstrates the steps performed in the feature engineering pipeline of the quantiles approach.

First, for each patient sample, we extracted values of the 7 vital sign features. Second, for each vital sign feature within that patient sample, we calculated the mean and SD. Third, we normalized the observations within each vital sign feature using the probability density function, and by passing the mean and SD calculated in step 2 as parameters to that function. The blue histograms in Figures 4 and 5 show the distribution of each vital sign feature before normalization, and the red curves show the distribution after normalization.

Figure 3. Feature engineering pipeline in the quantiles approach. MIMIC: Medical Information Mart for Intensive Care; ICU: intensive care unit; PDF: probability density function; PPF: percent point function.

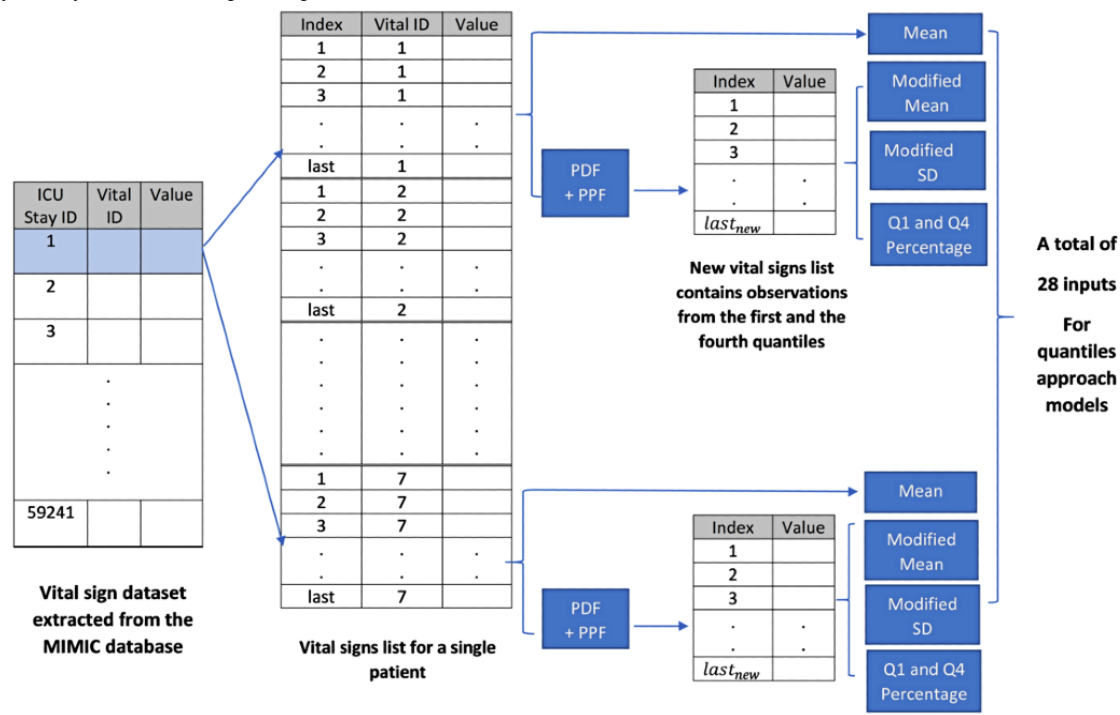

Fourth, we applied the percent point function (PPF) to each normalized vital sign feature to calculate two discrete values corresponding to the low and high values of that feature. The low values correspond to observations of the feature that are less than a given probability (the 25th percentile in our case) and the high values correspond to observations of the feature that are greater than or equal to a given probability (the 75th percentile in our case). Thus, for each vital sign feature, we calculated the values at which each percentage occurs.

Fifth, we used the calculated low and high values from step 4 to extract the observations of the vital sign features that occur in only the first and fourth quantiles (ie, we ignored the second and third quantiles). Sixth, we calculated the mean and SD of the extracted observations. In the remainder of the paper, we refer to these metrics as the modified mean and modified SD to distinguish from the original mean and SD calculated in step 2.

The final step is to calculate the quantile percentage for the vital sign feature by dividing the number of observations extracted in step 5 (ie, those that occur in the first and fourth quantiles) by the original number of observations (in all quantiles in the entire patient sample). Note that since we normalized the observations in the vital sign feature (step 3), the number of observations in the first and fourth quantiles will vary and will not always be 50% of the original observations.

**Figure 4.** Distribution of a sample patient observation before and after applying the quantiles approach.

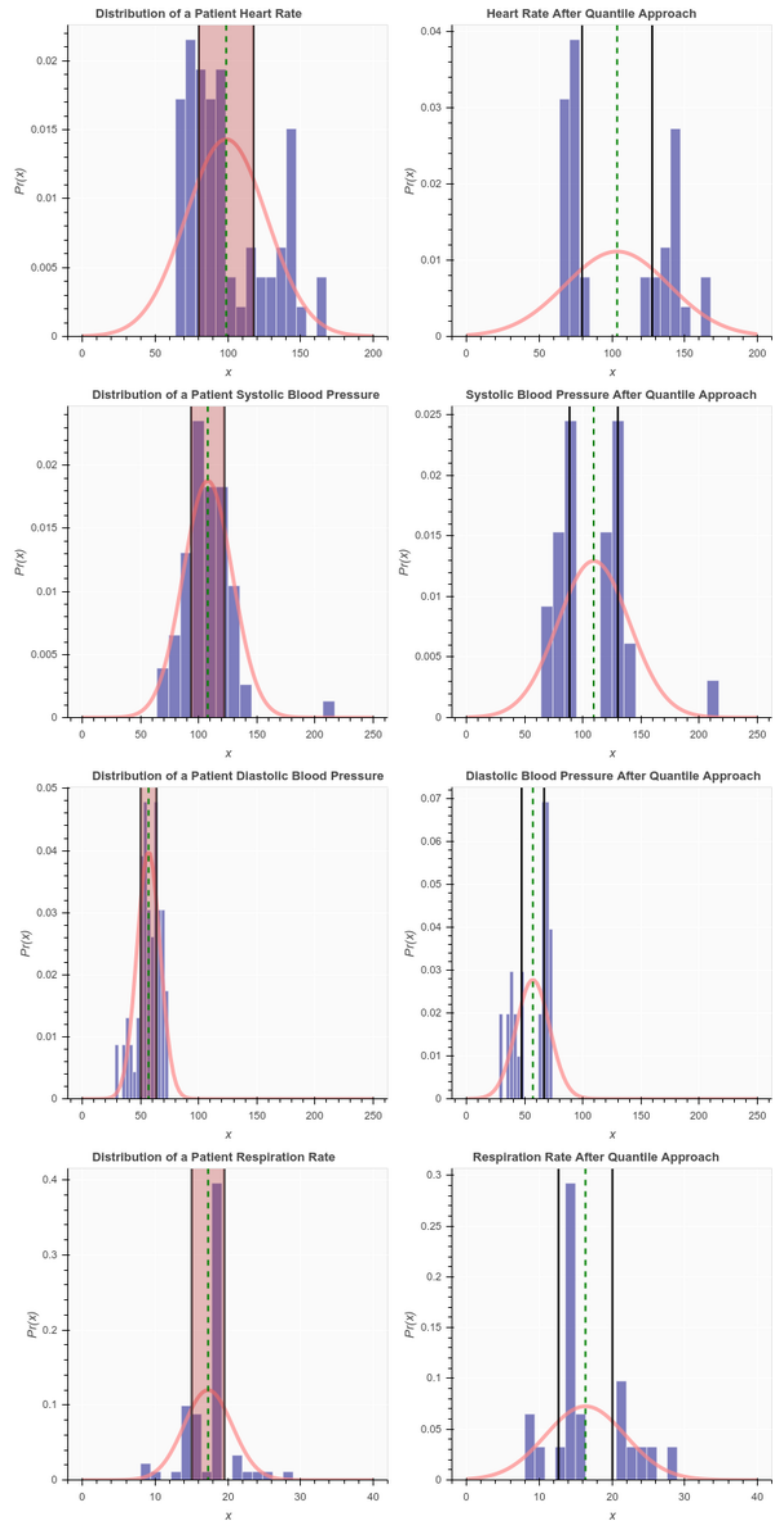

**Figure 5.** Distribution of a sample patient observation before and after applying the quantiles approach (continued from Figure 4).

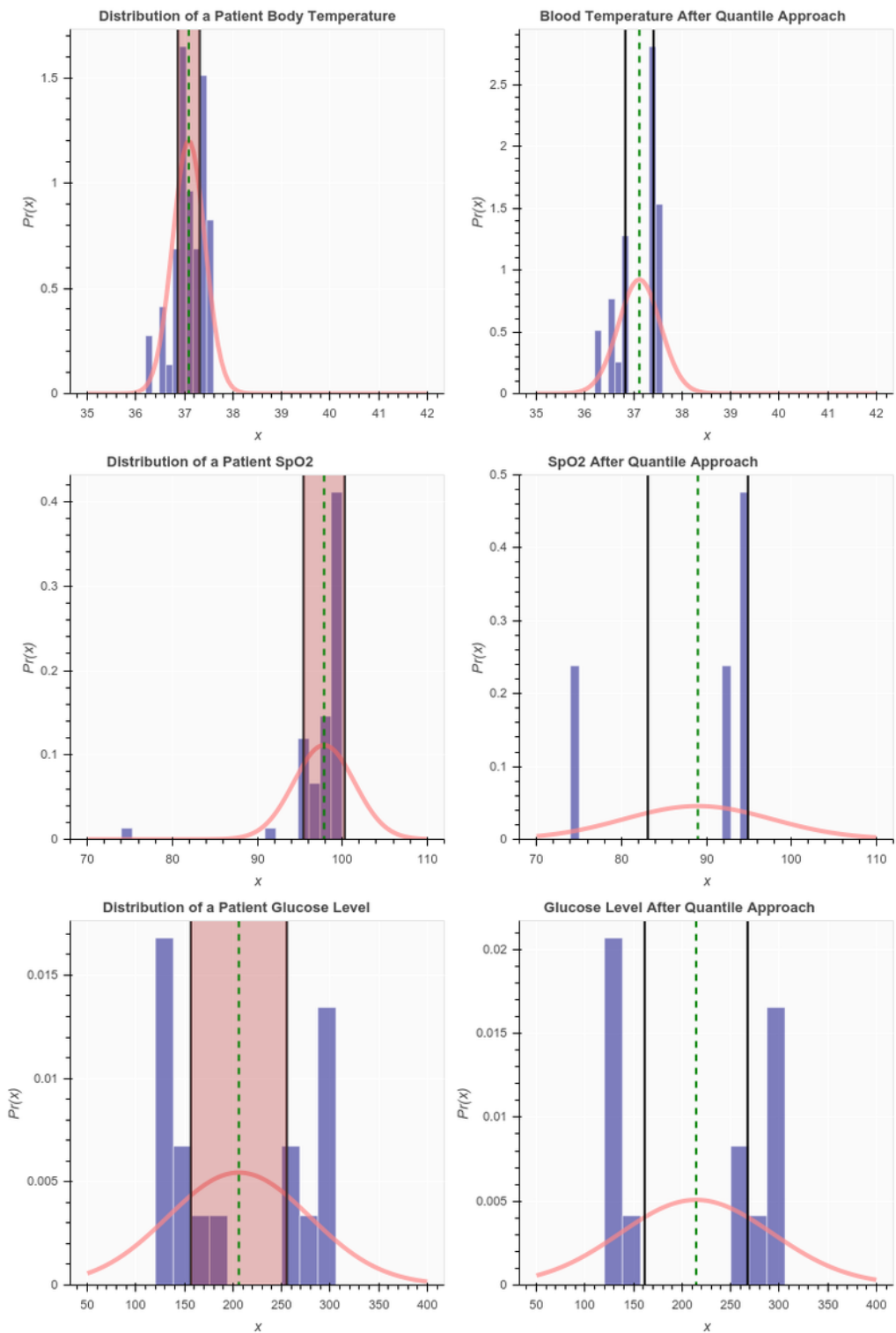

### Patient Use Case

To demonstrate the quantiles approach, we provide an example of a sample patient before and after applying the steps described above. Figures 4 and 5 show distributions of the 7 vital signs of the patient before (left) and (after) applying the quantiles approach. The shaded areas in Figures 4 and 5 show where the vital sign measurements are neglected. The right side of the figure shows the modified patient's observation after removing the values in the shaded area. After applying the change, the SD of the observation increased most of the time, whereas the mean (the green vertical line) did not change significantly. Table 4 shows an example of individual patient data before applying the quantiles approach. Table 5 demonstrates the features that were engineered from the original 7 vital sign measures for that patient sample.

**Table 4.** Sample data from an individual patient before applying the quantiles approach.

| Feature | Operationalization | Mean (SD) |
| --- | --- | --- |
| HeartRate_mean | Mean heart rate (beats/minute) | 98.92 (27.89) |
| sysbp_mean | Mean systolic blood pressure (mmHg) | 107.8 (21.26) |
| diasbp_mean | Mean diastolic blood pressure (mmHg) | 56.88 (10.00) |
| resprate_mean | Mean respiration rate (breaths/minute) | 17.29 (3.33) |
| tempc_mean | Mean body temperature (°C) | 37.08 (0.33) |
| spo$_2$_mean | Mean oxygen saturation (%) | 97.86 (3.57) |
| glucose_mean | Mean glucose level (mg/dL) | 206.0 (73.26) |

**Table 5.** Sample of patient data from after applying the quantiles approach.

| Feature | Operationalization | Value |
| --- | --- | --- |
| **Modified mean** | | |
| HeartRate_mean_mod | Mean of modified heart rate (beats/minute) | 103.59 |
| sysbp_mean_mod | Mean of modified arterial diastolic blood pressure (mmHg) | 109.34 |
| diasbp_mean_mod | Mean of modified arterial systolic blood pressure (mmHg) | 57.03 |
| resprate_mean_mod | Mean of modified respiratory rate (breaths/minute) | 16.36 |
| tempc_mean_mod | Mean of modified body temperature (°C) | 37.12 |
| spo$_2$_mean_mod | Mean of modified peripheral oxygen saturation (%) | 89.00 |
| glucose_mean_mod | Mean of modified blood glucose level (mg/dL) | 214.46 |
| **Modified SD** | | |
| heartRate_std_mod | SD of modified heart rate (beats/minute) | 35.76 |
| sysbp_std_mod | SD of modified arterial diastolic blood pressure (mmHg) | 30.83 |
| diasbp_std_mod | SD of modified arterial systolic blood pressure (mmHg) | 14.36 |
| resprate_std_mod | SD of modified respiratory rate (breaths/minute) | 5.49 |
| tempc_std_mod | SD of modified body temperature (°C) | 0.43 |
| spo$_2$_std_mod | SD of modified peripheral oxygen saturation (%) | 8.74 |
| glucose_std_mod | SD of modified blood glucose level (mg/dL) | 78.60 |
| **Modified quantiles** | | |
| HeartRateQuantPer | First and fourth quantiles percent of heart Rate | 0.5522 |
| SystolicQuantPer | First and fourth quantiles percent of arterial diastolic blood pressure | 0.4266 |
| DiastolicQuantPer | First and fourth quantiles percent of arterial systolic blood pressure | 0.4400 |
| RespRateQuantPer | First and fourth quantiles percent of respiratory rate | 0.3384 |
| TempCQuantPer | First and fourth quantiles percent of body temperature | 0.5384 |
| SPO$_2$QuantPer | First and fourth quantiles percent of peripheral oxygen saturation | 0.0689 |
| GlucoseQuantPer | First and fourth quantiles percent of blood glucose level | 0.8125 |

Table 6 lists additional features that were engineered from the original 7 vital sign measures using the quantiles approach for the entire patient population.

Pearson correlation analysis among the means of vital signs samples after applying the quantiles approach (Table 7) showed that there was no significant difference compared to the baseline model (Table 3). This implies that the quantiles approach does not considerably change the correlation between the variables.

**Table 6.** Vital sign data after applying the quantiles approach for the entire patient population.

| Feature | Operationalization | Value |
| --- | --- | --- |
| **Modified mean, mean (SD)** | | |
| HeartRate_mean_mod | Mean of modified heart rate (beats/minute) | 86.55 (15.8469) |
| sysbp_mean_mod | Mean of modified arterial diastolic blood pressure (mmHg) | 119.06 (16.865) |
| diasbp_mean_mod | Mean of modified arterial systolic blood pressure (mmHg) | 61.2201 (11.4944) |
| resprate_mean_mod | Mean of modified respiratory rate (breaths/minute) | 19.22 (4.1363) |
| tempc_mean_mod | Mean of modified body temperature (°C) | 36.82 (0.67382) |
| spo$_2$_mean_mod | Mean of modified peripheral oxygen saturation (%) | 96.00 (5.28098) |
| glucose_mean_mod | Mean of modified blood glucose level (mg/dL) | 144.50 (48.3843) |
| **Modified SD, mean (SD)** | | |
| heartRate_std_mod | SD of modified heart rate (beats/minute) | 11.33 (6.02761) |
| sysbp_std_mod | SD of modified arterial diastolic blood pressure (mmHg) | 19.22 (7.64726) |
| diasbp_std_mod | SD of modified arterial systolic blood pressure (mmHg) | 13.21 (6.06014) |
| resprate_std_mod | SD of modified respiratory rate (breaths/minute) | 4.96 (2.05444) |
| tempc_std_mod | SD of modified body temperature (°C) | 0.61 (0.35567) |
| spo$_2$_std_mod | SD of modified peripheral oxygen saturation (%) | 2.53 (2.18251) |
| glucose_std_mod | SD of modified blood glucose level (mg/dL) | 34.69 (32.2924) |
| **Modified quantiles, quantile percentage** | | |
| HeartRateQuantPer | First and fourth quantiles percent of heart Rate | 51.63 |
| SystolicQuantPer | First and fourth quantiles percent of arterial diastolic blood pressure | 50.49 |
| DiastolicQuantPer | First and fourth quantiles percent of arterial systolic blood pressure | 47.47 |
| RespRateQuantPer | First and fourth quantiles percent of respiratory rate | 49.02 |
| TempCQuantPer | First and fourth quantiles percent of body temperature | 56.57 |
| SPO$_2$QuantPer | First and fourth quantiles percent of peripheral oxygen saturation | 46.26 |
| GlucoseQuantPer | First and fourth quantiles percent of blood glucose level | 57.04 |

**Table 7.** Pearson correlation coefficients among the mean vital signs for a sample patient using the statistical model.

| Variable | Heart rate | Systolic BP[a] | Diastolic BP | Respiration rate | Body temperature | SPO$_2$[b] | Glucose |
| --- | --- | --- | --- | --- | --- | --- | --- |
| Heart rate | 1 | –0.103 | 0.183 | 0.316 | 0.236 | –0.065 | 0.053 |
| Systolic BP | –0.103 | 1 | 0.504 | –0.034 | 0.057 | 0.056 | 0.069 |
| Diastolic BP | 0.183 | 0.504 | 1 | 0.030 | 0.031 | 0.028 | 0.029 |
| Respiration rate | 0.316 | –0.034 | 0.030 | 1 | 0.128 | –0.095 | 0.064 |
| Body temperature | 0.236 | 0.057 | 0.031 | 0.128 | 1 | 0.016 | –0.028 |
| SPO$_2$ | –0.065 | 0.056 | 0.028 | –0.095 | 0.016 | 1 | –0.028 |
| Glucose | 0.053 | 0.069 | 0.029 | 0.064 | –0.028 | –0.028 | 1 |

[a]BP: blood pressure.
[b]SpO$_2$: oxygen saturation.

### Inputs to the Baseline Approach Versus the Quantiles Approach

The models built using the baseline approach used 12 predictor variables (ie, 5 demographic attributes and 7 vital signs) (Table 2). The feature engineering step performed in the quantiles approach augmented the original set of vital sign features with 21 extra features (ie, 7 variables corresponding to the mean of each patient observation, 7 variables corresponding to the SD of each patient observation, and 7 variables corresponding to the quantile percentages). Thus, in addition to the original 12 variables used in the baseline, the models built through the quantiles approach used the 21 engineered features.

## Classification Methods

### Models Applied

We used supervised learning techniques in both models for both variations because the model outputs were labeled accordingly. We split the dataset randomly into 75% as the training set (n=33,469 ICU stays) and 25% as the test set (n=11,157 ICU stays). To avoid overfitting, we used 10-fold cross-validation on the training set. We then trained both models using the training set and we validated the performance of both models using an unseen testing set.

We applied six commonly used ML algorithms for binary classification in both the mortality and LOS models: linear regression (LR), linear discriminant analysis, random forest (RF), k-nearest neighbors (kNN), support vector machine (SVM), and extreme gradient boosting (XGB). For the regression-based classification in the LOS model, we applied two ML algorithms to predict the number of days: MLR and support vector regression (SVR).

RF is an ensemble ML algorithm that generates bootstrapped samples from a dataset and uses the generated samples to construct several decision trees. Majority voting is then performed to decide the best classification of the generated samples. To avoid high correlation between the constructed trees, the algorithm uses a random subset of features to decide at each split point. This feature randomness increases the chances of having correct prediction results. Thus, one important parameter required by the algorithm is the number of features considered. In addition, choosing a high number of trees might increase the execution time with no considerable performance gain [21]. Therefore, another important parameter is the number of decision trees needed to compose the RF.

### Parameter Tuning for Mortality Classifiers

For the RF algorithm, we set the maximum number of features to consider for finding a good split to 4, and we set the estimated number of trees in an RF to 500. For SVM, we used the radial basis function as a kernel type and we set the penalty parameter of error *C* to 1.60.

### Parameter Tuning for LOS Classifiers

For the RF algorithm, we set the maximum number of features to consider in finding a good split to 4. We also set the estimated number of trees in the RF to 400. For SVM, we used the radial basis function as a kernel type and we set the penalty parameter of error *C* to 0.90.

## Model Calibration

To assess the goodness of fit in our models, we compared the accuracy on the test set and the mean accuracy of the trained model. We also used five metrics (accuracy, sensitivity, specificity, negative predictive value, and positive predictive value, along with corresponding 95% CIs) to validate the classification models on an unseen test set from the same population. We examined the difference in AUROC values between the test and training sets. Finally, we examined calibration across deciles using the sigmoid test supported with a visual inspection of calibration curves.

# Results

## Mortality Prediction Model

Table 8 shows the performance of the mortality models on both the training and test sets using the baseline and the quantiles approach with the six different ML algorithms.

The RF algorithm achieved the highest accuracy (88.61%) in predicting mortality on the test set using the quantiles approach, followed by the XGB algorithm with an accuracy of 88.22%. All models showed high specificity and low sensitivity, indicating that our models performed very well at identifying patients who will survive but not the opposite. XGB showed the highest sensitivity rate (0.16), demonstrating the advantage of using the XGB algorithm to identify patients who will not survive.

We observed relatively low improvement in model accuracy from the baseline approach to the quantiles approach. This can be explained by the imbalanced classification problem in the mortality model (ie, a low mortality rate of 11.89%). Another possible reason is that the sample size was reduced after applying the quantiles approach, which might have misled the classifier. The original sample size (44,626 ICU stays considering only the first day in the ICU) dropped by almost by half since we included only the first and fourth quantiles for each patient observation. The algorithm uses the PPF function to return discrete values that are less than or equal to the given probability, and the best probabilities achieved in our case were at the 25th and 75th percentiles. We tried other probabilities, but due to the small sample size, varying the PPF percent did not have a significant improvement on the results. Figure 6 shows a visual comparison between the accuracy of the six ML algorithms in the mortality model using the quantiles approach. The box plots to the left show the model accuracy on the training set using 10-fold cross-validation and the graph on the right shows the one-time model accuracy on the testing set.

**Table 8.** Mortality model results for six algorithms using different performance metrics.

| Method and algorithm | Training set accuracy, mean (SD) | Test set accuracy (95% CI) | Test set sensitivity (95% CI) | Test set specificity (95% CI) | Test set NPV[a] (95% CI) | Test set PPV[b] (95% CI) |
|---|---|---|---|---|---|---|
| **Baseline approach** | | | | | | |
| LR[c] | 0.8826 (0.0058) | 0.8806 (0.874-0.890) | 0.0331 (0.033-0.034) | 0.9979 (0.991-1.009) | 0.8817 (0.875-0.891) | 0.6923 (0.688-0.700) |
| LDA[d] | 0.8817 (0.0058) | 0.8788 (0.873-0.888) | 0.0523 (0.052-0.053) | 0.9932 (0.986-1.004) | 0.8833 (0.877-0.893) | 0.5182 (0.515-0.524) |
| RF[e] | 0.8846 (0.0061) | 0.8854 (0.879-0.895) | 0.1127 (0.112-0.114) | 0.9923 (0.985-1.003) | 0.8898 (0.884-0.899) | 0.6710 (0.666-0.679) |
| kNN[f] | 0.8765 (0.0054) | 0.8760 (0.870-0.886) | 0.0854 (0.085-0.087) | 0.9855 (0.978-0.996) | 0.8861 (0.880-0.896) | 0.4496 (0.447-0.455) |
| SVM[g] | 0.8837 (0.0058) | 0.8808 (0.875-0.890) | 0.0272 (0.027-0.028) | 0.9989 (0.992-1.010) | 0.8811 (0.875-0.891) | 0.7872 (0.782-0.796) |
| XGB[h] | 0.8842 (0.0061) | 0.8815 (0.875-0.891) | 0.1429 (0.142-0.145) | 0.9837 (0.977-0.994) | 0.8923 (0.886-0.902) | 0.5495 (0.546-0.556) |
| **Quantiles approach** | | | | | | |
| LR | 0.8838 (0.0063) | 0.8815 (0.875-0.891) | 0.0545 (0.054-0.055) | 0.9960 (0.989-1.007) | 0.8838 (0.878-0.893) | 0.6548 (0.650-0.662) |
| LDA | 0.8821 (0.0067) | 0.8814 (0.875-0.891) | 0.0935 (0.093-0.095) | 0.9905 (0.983-1.001) | 0.8875 (0.881-0.897) | 0.5772 (0.573-0.584) |
| RF | 0.8859 (0.0064) | 0.8861 (0.880-0.896) | 0.0891 (0.089-0.090) | 0.9964 (0.989-1.007) | 0.8876 (0.881-0.897) | 0.7756 (0.770-0.784) |
| KNN | 0.8802 (0.0060) | 0.8764 (0.870-0.886) | 0.0589 (0.059-0.060) | 0.9895 (0.982-1.000) | 0.8836 (0.877-0.893) | 0.4395 (0.437-0.445) |
| SVM | 0.8851 (0.0058) | 0.8820 (0.876-0.892) | 0.0449 (0.045-0.046) | 0.9816 (0.991-1.009) | 0.8829 (0.877-0.893) | 0.7439 (0.739-0.752) |
| XGB | 0.8844 (0.0061) | 0.8822 (0.875-0.891) | 0.1643 (0.164-0.167) | 0.9816 (0.975-0.992) | 0.8945 (0.888-0.904) | 0.5533 (0.550-0.560) |

[a]NPV: negative predictive value.
[b]PPV: positive predictive value.
[c]LR: logistic regression.
[d]LDA: linear discriminant analysis.
[e]RF: random forest.
[f]kNN: k-nearest neighbor.
[g]SVM: support vector machine.
[h]XGB: extreme gradient boosting.

**Figure 6.** Comparison of the mortality model results using the quantiles approach on the training set (left) and the test set (right). LR: logistic regression; LDA: linear discriminant analysis; RF: random forest; KNN: k-nearest neighbor; SVM: support vector machine; XGB: extreme gradient boosting.

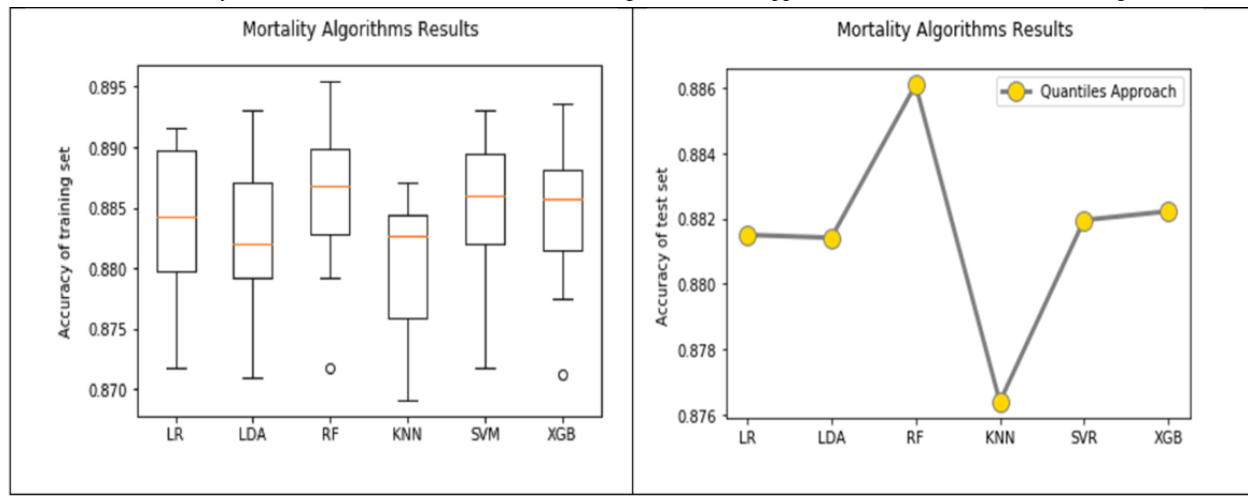

The ROC curve is commonly used to evaluate the performance of an ML model by showing the relationship between the false-positive and true-positive rates. The AUROC metric can be used as a basis for comparison; higher values indicate that a model can identify classes using a specific ML algorithm better than another. In the case of the mortality model, the ROC curve shows the relationship between survival cases that scored as no survival and no survival cases that scored as no survival.

Table 9 shows the AUROC results of the mortality model on both the training and test sets using the baseline and quantile approaches for the different ML algorithms. Figure 7 shows the ROC curves for the six ML algorithms for both the baseline and the quantiles approach. XGB produced the highest AUROC (0.79) for predicting mortality on the test set using the quantiles approach (Table 9).

**Table 9.** Mortality model performance based on area under the receiver operating characteristic curve (AUROC).

| Method and algorithm | Training set AUROC, mean (SD) | Test set AUROC |
| --- | --- | --- |
| **Baseline approach** | | |
| LR[a] | 0.702047 (0.015652) | 0.69313 |
| LDA[b] | 0.701731 (0.016077) | 0.69247 |
| RF[c] | 0.764875 (0.009214) | 0.76725 |
| kNN[d] | 0.629262 (0.008944) | 0.63173 |
| SVM[e] | 0.653269 (0.011730) | 0.66800 |
| XGB[f] | 0.771187 (0.012094) | 0.76971 |
| **Quantiles approach** | | |
| LR | 0.727331 (0.014217) | 0.72810 |
| LDA | 0.725909 (0.014758) | 0.72622 |
| RF | 0.783696 (0.010503) | 0.78292 |
| KNN | 0.631649 (0.010416) | 0.64087 |
| SVM | 0.719253 (0.008940) | 0.72333 |
| XGB | 0.788908 (0.010665) | 0.79036 |

[a]LR: logistic regression.

[b]LDA: linear discriminant analysis.

[c]RF: random forest.

[d]kNN: k-nearest neighbor.

[e]SVM: support vector machine.

[f]XGB: extreme gradient boosting.

**Figure 7.** Comparison of receiver operating characteristic curves in the mortality model using the baseline (left) and the quantiles approach (right). LR: logistic regression; LDA: linear discriminant analysis; RF: random forest; KNN: k-nearest neighbour; SVM: support vector machine; XGB: extreme gradient boosting.

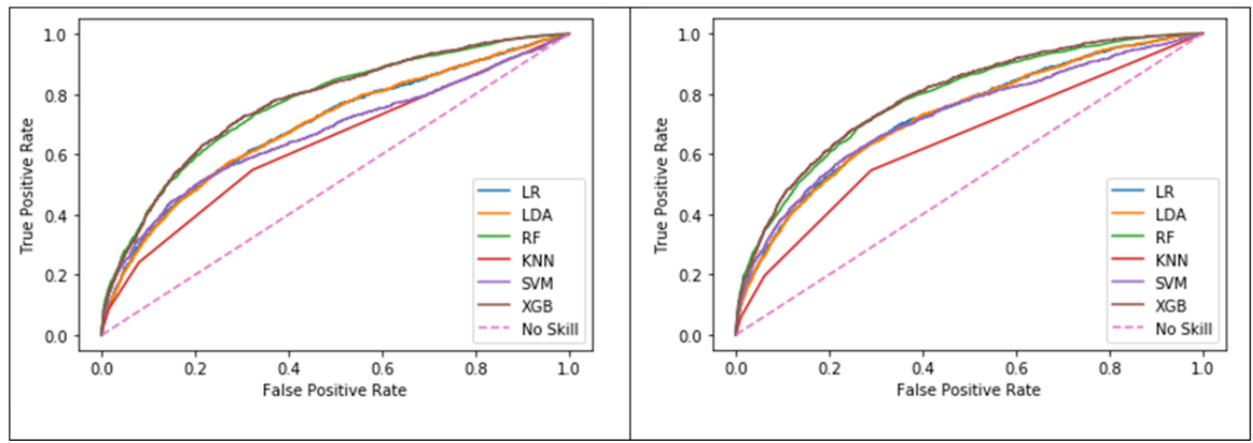

## LOS Prediction Model

### Binary Classification Algorithms

Table 10 shows the performance of the binary classification models for the LOS model on both the training set and test set using the baseline and the quantiles approaches with the six different ML algorithms.

**Table 10.** Length of stay model results for six algorithms using different performance metrics.

| Method and algorithm | Training set accuracy, mean (SD) | Test set accuracy (95% CI) | Test set sensitivity (95% CI) | Test set specificity (95% CI) | Test set NPV[a] (95% CI) | Test set PPV[b] (95% CI) |
|---|---|---|---|---|---|---|
| **Baseline approach** | | | | | | |
| LR[c] | 0.5787 (0.01) | 0.5715 (0.57-0.58) | 0.56 (0.554-0.564) | 0.59 (0.58-0.59) | 0.57 (0.563-0.573) | 0.58 (0.57-0.58) |
| LDA[d] | 0.5787 (0.01) | 0.5710 (0.57-0.58) | 0.56 (0.55-0.56) | 0.59 (0.58-0.59) | 0.57 (0.56-0.57) | 0.58 (0.57-0.58) |
| RF[e] | 0.6205 (0.01) | 0.6193 (0.62-0.63) | 0.61 (0.60-0.61) | 0.63 (0.63-0.64) | 0.61 (0.61-0.62) | 0.63 (0.62-0.63) |
| kNN[f] | 0.5639 (0.01) | 0.5713 (0.57-0.58) | 0.52 (0.520-0.529) | 0.62 (0.616-0.627) | 0.56 (0.559-0.569) | 0.58 (0.58-0.59) |
| SVM[g] | 0.6228 (0.01) | 0.6141 (0.61-0.62) | 0.56 (0.56-0.57) | 0.67 (0.66-0.68) | 0.60 (0.60-0.61) | 0.63 (0.63-0.64) |
| XGB[h] | 0.6303 (0.01) | 0.6130 (0.61-0.62) | 0.58 (0.58-0.59) | 0.64 (0.64-0.65) | 0.60 (0.60-0.61) | 0.62 (0.62-0.63) |
| **Quantiles approach** | | | | | | |
| LR | 0.6126 (0.01) | 0.6131 (0.61-0.62) | 0.59 (0.59-0.60) | 0.63 (0.629-0.640) | 0.61 (0.60-0.61) | 0.62 (0.62-0.63) |
| LDA | 0.6131 (0.01) | 0.6130 (0.61-0.62) | 0.59 (0.59-0.60) | 0.64 (0.63-0.64) | 0.61 (0.60-0.61) | 0.62 (0.62-0.63) |
| RF | 0.6511 (0.01) | 0.6461 (0.64-0.65) | 0.64 (0.63-0.66) | 0.66 (0.65-0.66) | 0.64 (0.64-0.65) | 0.65 (0.65-0.66) |
| kNN | 0.5748 (0.01) | 0.5768 (0.57-0.58) | 0.4865 (0.483-0.49) | 0.6681 (0.66-0.68) | 0.56 (0.56-0.57) | 0.60 (0.59-0.60) |
| SVM | 0.6466 (0.01) | 0.6386 (0.63-0.65) | 0.5939 (0.59-0.60) | 0.68 (0.68-0.69) | 0.63 (0.62-0.63) | 0.66 (0.65-0.66) |
| XGB | 0.6496 (0.01) | 0.6284 (0.62-0.64) | 0.61 (0.60-0.62) | 0.65 (0.64-0.66) | 0.62 (0.62-0.63) | 0.64 (0.63-0.64) |

[a]NPV: negative predictive value.
[b]PPV: positive predictive value.
[c]LR: logistic regression.
[d]LDA: linear discriminant analysis.
[e]RF: random forest.
[f]kNN: k-nearest neighbor.
[g]SVM: support vector machine.
[h]XGB: extreme gradient boosting.

The best accuracy of predicting ICU LOS on the test set was 64.64% using the RF algorithm in the quantiles approach, followed by the SVM algorithm with an accuracy of 63.86%. The improvement in model accuracy from the baseline approach to the quantiles approach was better when compared with that found for the mortality model (Table 8). For example, the difference in accuracy between the baseline and the quantiles approach for the LOS model on the test set was 2.68% using RF and was 2.45% using SVM. The RF algorithm achieved the highest sensitivity (0.64), which indicates that the model using the RF algorithm can identify patients who will stay in the ICU for more than 2.64 days better than the other algorithms. SVM achieved the highest specificity (0.68), which indicates that the model using the SVM algorithm is better at identifying patients who will stay in the ICU for 2.64 days or less compared with the other algorithms. Figure 8 shows a visual comparison of the accuracy of the six algorithms in the LOS model results using the quantiles approach. The box plots on the left show the model accuracy on the training set using 10-fold cross-validation and the graph on the right shows the one-time model accuracy on the testing set.

Table 11 shows the AUROC results of the LOS model on both the training and test sets using the baseline and the quantiles approach with the six ML algorithms. Figure 9 shows the ROC curves for the algorithms in the baseline approach and the quantiles approach, respectively. The RF algorithm using the quantiles approach produced the highest AUROC (0.697) for predicting the LOS on the test set (Table 11).

**Figure 8.** Comparison of the length of stay model results using the quantiles approach on the training set (left) and the test set (right). LR: logistic regression; LDA: linear discriminant analysis; RF: random forest; KNN: k-nearest neighbor; SVM: support vector machine; XGB: extreme gradient boosting.

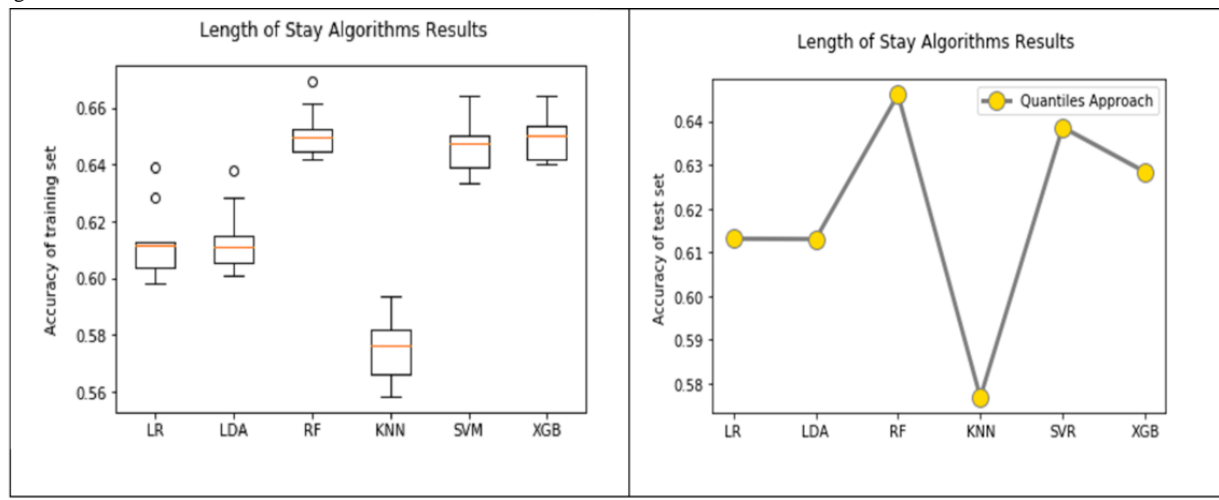

**Table 11.** Performance of the length of stay model results based on the area under the receiver operating characteristic curve (AUROC).

| Method and algorithm | Training set AUROC, mean (SD) | Test set AUROC |
| --- | --- | --- |
| **Baseline approach** | | |
| LR[a] | 0.612883 (0.006047) | 0.60833 |
| LDA[b] | 0.612776 (0.006058) | 0.60837 |
| RF[c] | 0.664959 (0.006147) | 0.66325 |
| kNN[d] | 0.583710 (0.006401) | 0.59110 |
| SVM[e] | 0.665992 (0.006041) | 0.66118 |
| XGB[f] | 0.677454 (0.007311) | 0.66586 |
| **Quantiles approach** | | |
| LR | 0.654390 (0.012180) | 0.65407 |
| LDA | 0.654178 (0.012102) | 0.65384 |
| RF | 0.705115 (0.010004) | 0.69782 |
| kNN | 0.598228 (0.007539) | 0.60507 |
| SVM | 0.694473 (0.009834) | 0.69272 |
| XGB | 0.704889 (0.011338) | 0.69693 |

[a]LR: logistic regression.
[b]LDA: linear discriminant analysis.
[c]RF: random forest.
[d]kNN: k-nearest neighbor.
[e]SVM: support vector machine.
[f]XGB: extreme gradient boosting.

**Figure 9.** Comparison of receiver operating characteristic curves in the length of stay model using the baseline (left) and quantiles (right) approaches. LR: logistic regression; LDA: linear discriminant analysis; RF: random forest; KNN: k-nearest neighbor; SVM: support vector machine; XGB: extreme gradient boosting.

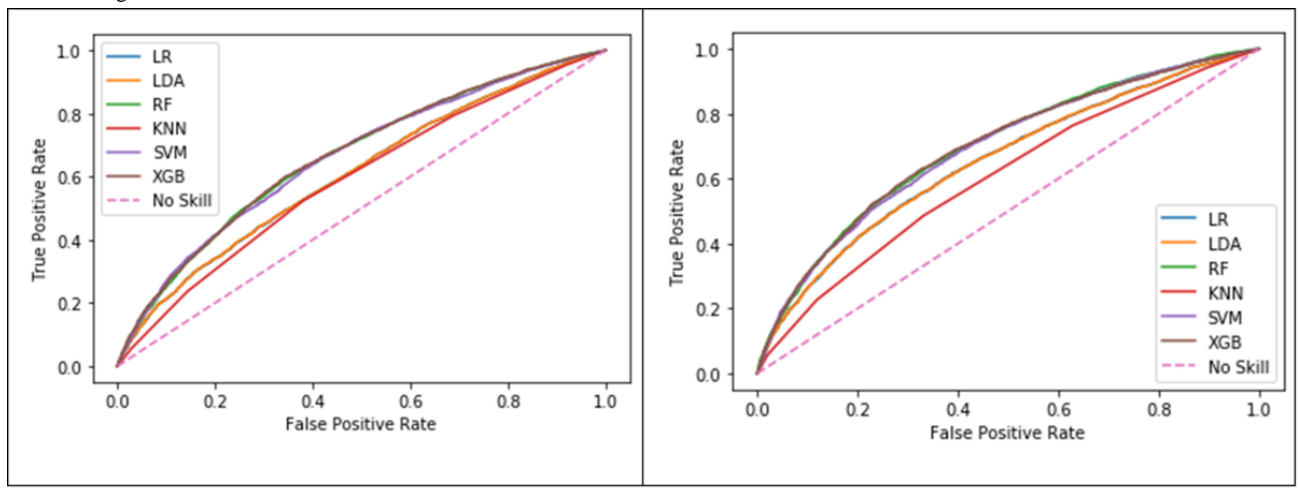

### Regression-Based Classifiers

As for the regression-based classifiers of the LOS model, we report the error between the predicted values and actual values in the test set using both the mean absolute error (MAE) and the root mean squared error metrics. The minimum, mean, and maximum LOS for the entire population was 1, 2.64, and 173.07 days, respectively. Table 12 shows the error value (per day) using both error metrics for the LOS model. The lowest error value obtained was 2.81 days using the MAE in the SVR algorithm with the quantiles approach.

Table 12. Regression error values of the length of stay model using the baseline and quantile approaches.

| Method | MAE[a] | RMSE[b] |
| --- | --- | --- |
| **Baseline approach** | | |
|     MLR[c] | 3.509 | 6.029 |
|     SVR[d] | 2.857 | 6.214 |
| **Quantiles approach** | | |
|     MLR | 3.446 | 5.961 |
|     SVR | 2.810 | 6.137 |

[a]MAE: mean absolute error.
[b]RMSE: root mean square error.
[c]MLR: multivariate linear regression.
[d]SVR: support vector regression.

## Discussion

### Principal Results

Our findings indicate that we can build prediction models for ICU LOS and mortality with better accuracy using a combination of ML and the quantiles approach including only vital signs. Little improvement in the accuracy of the mortality model was achieved, but improvement of approximately 2.7% was achieved in the LOS model using the proposed quantiles approach. We examined model calibration across deciles for all six algorithms in both models. Figure 10 shows the probability calibration curves of the mortality model using the six algorithms. The six plots show good calibration of the models, especially in the case of the RF algorithm. Figure 11 shows the probability calibration curves of the LOS model using the six algorithms. The six plots show good calibration of the models except for the kNN algorithm.

Figure 10. Probability calibration curves of the mortality model for the six classification algorithms. LR: logistic regression; LDA: linear discriminant analysis; RF: random forest; KNN: k-nearest neighbor; SVM: support vector machine; XGB: extreme gradient boosting.

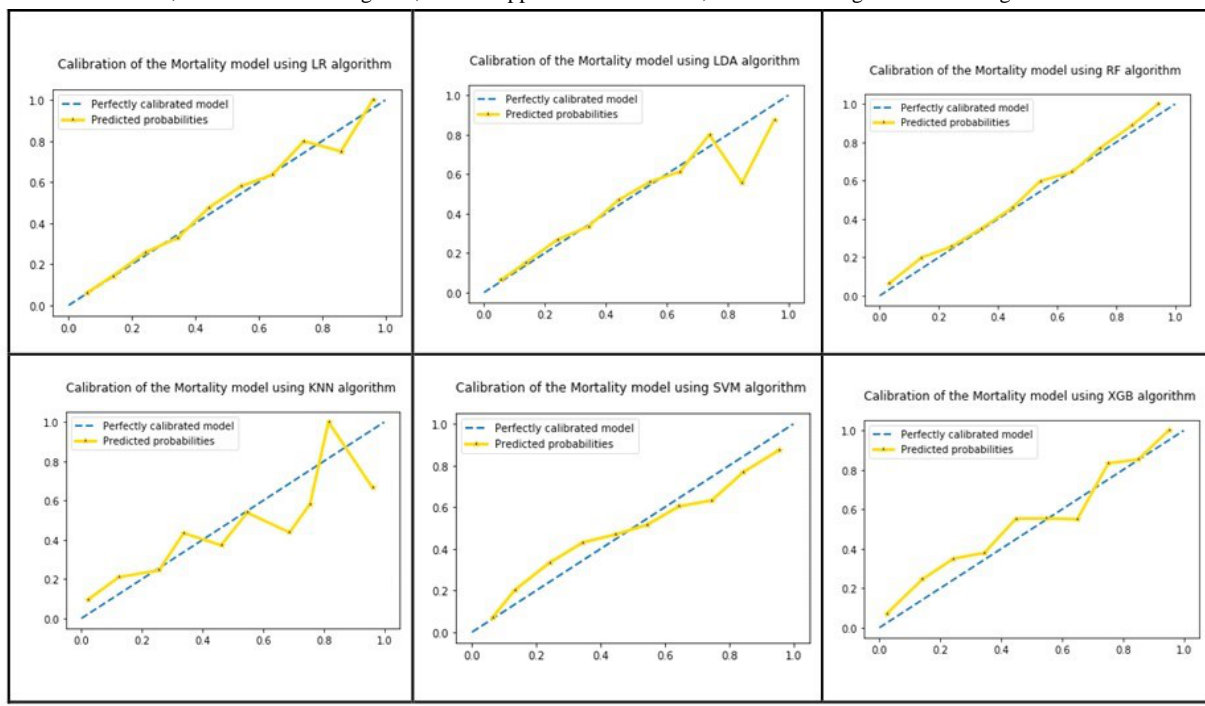

**Figure 11.** Probability calibration curves of the length of stay model for the six classification algorithms. LR: logistic regression; LDA: linear discriminant analysis; RF: random forest; KNN: k-nearest neighbor; SVM: support vector machine; XGB: extreme gradient boosting.

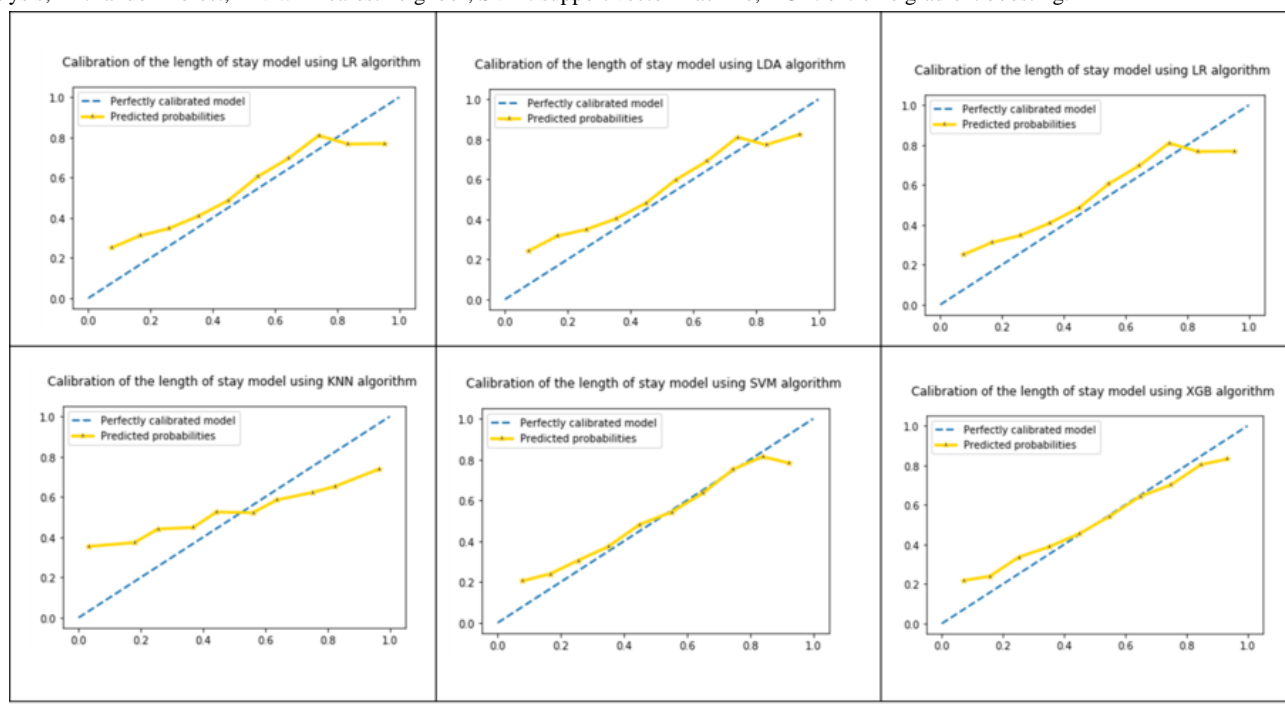

One might argue that we included only the mean and not the SD of the vital signs in the baseline approach when the comparison was to a model including both the mean and SD in the quantiles approach. Both the baseline and the quantile approaches include the means of vital signs. The quantiles approach includes an extra 21 features corresponding to modified means and modified SDs of the original values in addition to the quantile percentages. Had we chosen to include both the mean and SD of the original vital sign observations in the baseline approach, we would have also needed to include the SD of the original vital sign observations in the quantiles approach. In this case, we do not expect that there will be a significant impact.

Moreover, based on the method of population selection, the same patient could be in the training as well as in the test set but for different ICU admissions at different time points. For this study, we considered unique ICU admissions as opposed to unique patient identifiers. The rationale for focusing on unique admissions is that we sought to predict mortality and LOS without having prior knowledge about a patient's medical conditions or diagnoses.

## Qualitative Comparison With Other Approaches

For the mortality model, we were able to achieve approximately 89% accuracy and an AUROC of 0.78 using only 7 vital sign features and 4 demographic attributes, along with 21 features engineered from the original features. Other researchers have used excessively more features to achieve similar or better accuracies. For instance, Johnson et al [12] used a total of 148 features to achieve an AUROC of 0.92. Lehman et al [14] applied the SAPS-I algorithm on commonly used physiological data to predict mortality and achieved an AUROC of 0.72. Johnson et al [13] used a range of features, including standard statistical descriptors, to achieve an AUROC of 0.86.

For LOS models, most researchers used an exhaustive list of features to achieve higher accuracy in their models, but they did not report on whether they had balanced classification problems. For example, Harutyunyan et al [16] achieved 84% accuracy using 17 clinical variables and by considering a target ICU LOS of 7 days. Gentimis et al [17] achieved 80% accuracy using several inputs from seven tables to build the LOS model with a target ICU stay of 5 days. Bertsimas et al [18] used several static and dynamic variables, and achieved accuracy in the >80% range. In our approach, we built balanced classification models (using the median LOS of the entire population) with minimal features. These two conditions made it harder to achieve high accuracy, which reached only 65% in the LOS model.

One contribution of our method is the unique combination of ML with the quantiles approach. Other researchers have used various techniques to assess a patient's deteriorating conditions. Tyler et al [15] found that the methods to normalize patients' abnormal values are not thoroughly correct and might affect the results negatively. Other researchers relied on scoring systems (eg, centile-based early warning score, National Early Warning Score, or SAPS) to estimate or recognize patients' deteriorating conditions. We avoided relying on existing early warning scoring systems since they vary from patient to patient, which may lead to uncertain results.

## Sensitivity Analysis

Since we considered unique ICU stays rather than individual patients, the training set/testing set split was not performed at the patient level. This might raise the concern that the vital signs and LOS measured at different ICU visits for the same patient could be highly correlated. Thus, the mortality and the LOS models might risk overestimation in predictive performance. We mitigated this effect by performing a sensitivity analysis to compare the results of the models after excluding patient overlap to the results of the original model with the overlap included. In the original model, the population size was 44,626 (corresponding to ICU stays), the training set size was 33,469 ICU stays (75% of the population), and the test set size was 11,157 ICU stays (25% of the population).

The patient overlap between the training and test sets was 3886 ICU stays (34.83% of the test set). The number of ICU stays remaining in the test set after removing the patient overlap (ie, 3886) reduced to 7271 (65.17% of the original test set of size 11,157). Table 13 shows the results of the mortality model after removing the overlap and Table 14 shows the results of the LOS model after removing the overlap. There were no significant changes compared to the model results shown in Table 8 and Table 10, respectively.

**Table 13.** Mortality model results for six algorithms using different performance metrics.

| Methods and algorithm | Training set accuracy, mean (SD) | Test set accuracy (95% CI) | Test set sensitivity (95% CI) | Test set specificity (95% CI) | Test set NPV[a] (95% CI) | Test set PPV[b] (95% CI) |
|---|---|---|---|---|---|---|
| **Quantiles approach without overlap in the test set** | | | | | | |
| LR[c] | 0.88263 (0.0058) | 0.87636 (0.870-0.886) | 0.0620 (0.062-0.063) | 0.9963 (0.989-1.007) | 0.8781 (0.872-0.888) | 0.7160 (0.711-0.724) |
| LDA[d] | 0.88171 (0.0058) | 0.87663 (0.870-0.886) | 0.0974 (0.097-0.099) | 0.9914 (0.984-1.002) | 0.8817 (0.875-0.891) | 0.6275 (0.623-0.635) |
| RF[e] | 0.88458 (0.0061) | 0.88145 (0.875-0.891) | 0.0952 (0.095-0.097) | 0.9973 (0.990-1.008) | 0.8820 (0.876-0.892) | 0.8396 (0.834-0.849) |
| kNN[f] | 0.87645 (0.0054) | 0.87196 (0.866-0.881) | 0.0620 (0.062-0.063) | 0.9913 (0.984-1.002) | 0.8776 (0.871-0.887) | 0.5132 (0.510-0.519) |
| SVM[g] | 0.88365 (0.0058) | 0.87581 (0.870-0.885) | 0.0428 (0.042-0.044) | 0.9985 (0.991-1.009) | 0.8762 (0.875-0.891) | 0.8163 (0.811-0.825) |
| XGB[h] | 0.88422 (0.0061) | 0.87966 (0.873-0.889) | 0.1670 (0.166-0.169) | 0.9846 (0.978-0.995) | 0.8891 (0.883-0.899) | 0.6166 (0.612-0.624) |
| **Quantiles approach** | | | | | | |
| LR | 0.88380 (0.0063) | 0.88150 (0.875-0.891) | 0.0545 (0.054-0.055) | 0.9960 (0.989-1.007) | 0.8838 (0.878-0.893) | 0.6548 (0.650-0.662) |
| LDA | 0.88210 (0.0067) | 0.88141 (0.875-0.891) | 0.0935 (0.093-0.095) | 0.9905 (0.983-1.001) | 0.8875 (0.881-0.897) | 0.5772 (0.573-0.584) |
| RF | 0.88586 (0.0064) | 0.88608 (0.880-0.896) | 0.0891 (0.089-0.090) | 0.9964 (0.989-1.007) | 0.8876 (0.881-0.897) | 0.7756 (0.770-0.784) |
| KNN | 0.88018 (0.0060) | 0.87640 (0.870-0.886) | 0.0589 (0.059-0.060) | 0.9895 (0.982-1.000) | 0.8836 (0.877-0.893) | 0.4395 (0.437-0.445) |
| SVM | 0.88511 (0.0058) | 0.88195 (0.876-0.892) | 0.0449 (0.045-0.046) | 0.9816 (0.991-1.009) | 0.8829 (0.877-0.893) | 0.7439 (0.739-0.752) |
| XGB | 0.88443 (0.0061) | 0.88222 (0.875-0.891) | 0.1643 (0.164-0.167) | 0.9816 (0.975-0.992) | 0.8945 (0.888-0.904) | 0.5533 (0.550-0.560) |

[a]NPV: negative predictive value.
[b]PPV: positive predictive value.
[c]LR: logistic regression.
[d]LDA: linear discriminant analysis.
[e]RF: random forest.
[f]kNN: k-nearest neighbor.
[g]SVM: support vector machine.
[h]XGB: extreme gradient boosting.

**Table 14.** Length of stay model results for six algorithms using different performance metrics.

| Method and algorithm | Training set accuracy, mean (SD) | Test set accuracy (95% CI) | Test set sensitivity (95% CI) | Test set specificity (95% CI) | Test set NPV[a] (95% CI) | Test set PPV[b] (95% CI) |
|---|---|---|---|---|---|---|
| **Quantiles approach without overlap in the test set** | | | | | | |
| LR[c] | 0.61262 (0.0117) | 0.61312 (0.609-0.620) | 0.5983 (0.594-0.605) | 0.6267 (0.622-0.634) | 0.6292 (0.625-0.636) | 0.5957 (0.592-0.602) |
| LDA[d] | 61.307 (0.0112) | 0.61216 (0.608-0.619) | 0.5951 (0.591-0.602) | 0.6277 (0.624-0.635) | 0.6277 (0.624-0.635) | 0.5951 (0.591-0.602) |
| RF[e] | 0.65108 (0.0081) | 0.64778 (0.643-0.655) | 0.6400 (0.636-0.647) | 0.6550 (0.651-0.662) | 0.6643 (0.660-0.672) | 0.6304 (0.626-0.637) |
| kNN[f] | 0.57483 (0.0104) | 0.58740 (0.583-0.594) | 0.4941 (0.491-0.500) | 0.6731 (0.669-0.681) | 0.5913 (0.587-0.598) | 0.5816 (0.578-0.588) |
| SVM[g] | 0.64659 (0.0088) | 0.64379 (0.639-0.651) | 0.5946 (0.591-0.601) | 0.6890 (0.684-0.697) | 0.6489 (0.645-0.656) | 0.6374 (0.633-0.645) |
| XGB[h] | 0.64961 (0.0076) | 0.63540 (0.631-0.642) | 0.6132 (0.609-0.620) | 0.6557 (0.651-0.663) | 0.6483 (0.644-0.656) | 0.6209 (0.617-0.628) |
| **Quantiles approach** | | | | | | |
| LR | 0.61262 (0.0117) | 0.61307 (0.609-0.620) | 0.5930 (0.589-0.600) | 0.6332 (0.629-0.640) | 0.6058 (0.602-0.613) | 0.6208 (0.617-0.628) |
| LDA | 0.61307 (0.0112) | 0.61298 (0.609-0.620) | 0.5909 (0.587-0.598) | 0.6352 (0.631-0.642) | 0.6053 (0.601-0.612) | 0.6212 (0.617-0.628) |
| RF | 0.65108 (0.0081) | 0.64614 (0.642-0.653) | 0.6374 (0.633-0.645) | 0.6549 (0.650-0.662) | 0.6408 (0.636-0.648) | 0.6516 (0.647-0.659) |
| KNN | 0.57483 (0.0104) | 0.57677 (0.573-0.583) | 0.4865 (0.483-0.492) | 0.6681 (0.664-0.676) | 0.5624 (0.559-0.569) | 0.5974 (0.593-0.604) |
| SVM | 0.64659 (0.0088) | 0.63861 (0.634-0.646) | 0.5939 (0.590-0.601) | 0.6838 (0.679-0.691) | 0.6245 (0.620-0.632) | 0.6553 (0.651-0.663) |
| XGB | 0.64961 (0.0076) | 0.62839 (0.624-0.635) | 0.6085 (0.604-0.615) | 0.6484 (0.644-0.656) | 0.6206 (0.617-0.628) | 0.6367 (0.632-0.644) |

[a]NPV: negative predictive value.

[b]PPV: positive predictive value.

[c]LR: logistic regression.

[d]LDA: linear discriminant analysis.

[e]RF: random forest.

[f]kNN: k-nearest neighbor.

[g]SVM: support vector machine.

[h]XGB: extreme gradient boosting.

The total number of ICU stays was 44,626 and the total number of patients was 33,466. We calculated the frequency of ICU stays for the entire patient population. We found that 80% of the population visited the ICU only once and 20% visited the ICU more than once. Moreover, the MIMIC database includes data for patients who might have stayed in different ICU types (eg, general, cardiac) and due to different health conditions. In addition, a patient might have visited one ICU more frequently than another, and the time period between consecutive visits within a single ICU might be several years. The sensitivity analysis findings in our case might be due to the fact that our approach focused on the visits rather than the patients and ignored the details mentioned above.

### Limitations

Admittedly, this study lacks quantitative comparisons with previous research on the same topic owing to substantial differences between the research questions tackled previously, and the associated data extraction pipelines and assumptions. We mitigated this limitation by providing a qualitative comparison between our models and previous models.

Previous research based on data from the MIMIC database likely demonstrated higher accuracy since excessively more features were used than applied in this study. We believe that it is difficult to achieve high model accuracy using a limited number of features.

Additionally, as in any ML-based method, our approach might have some limitations. In this study, we used the MIMC database, which represents a patient population from a single hospital in Boston, and does not generalize to other populations or hospital systems in other areas across the United States or the rest of the world. Future research will focus on applying our approach to other patient populations.

Moreover, we ran the models using only the vital signs to measure the impact of the demographic attributes. We found that the effects of demographic attributes on the results were low. For example, age did not have a considerable effect since we were only using adult patient data in the MIMIC database. The accuracy of the mortality model without the age feature using RF in the quantiles approach was 88.536%, which is very close to the model result obtained when including age. The mortality model achieved an AUROC of 0.77 without using age and 0.78 with age included. The accuracy of the LOS model without including the age feature using RF and the quantiles approach was 64.39%, which is very close to the result obtained with the age feature included. Table 15 also shows that the differences in AUROC and positive predictive value were not significant between the mortality and LOS models both including and excluding the age feature using the RF algorithm and the quantiles approach. This would be different in pediatrics and adolescent populations, for whom vital measurements are more age-sensitive. In addition, in the MIMIC database, the ages for patients older than 89 years are not accurate; we used 90 years as a dummy value for all of these patients. Another potential reason for the low impact of including the demographic attributes is the lack of variation in height due to missing values that had to be imputed using the population mean.

Table 15. Model results including and excluding the age feature.

| Model | Accuracy | AUROC[a] | PPV[b] (95% CI) |
|---|---|---|---|
| **Mortality** | | | |
| Without age | 88.536 | 0.76740 | 0.7468 (0.742-0.755) |
| With age | 88.608 | 0.78292 | 0.7756 (0.770-0.784) |
| **Length of stay** | | | |
| Without age | 64.390 | 0.69433 | 0.6487 (0.644-0.656) |
| With age | 64.614 | 0.69782 | 0.6516 (0.647-0.659) |

[a]AUROC: area under the receiver operating characteristic curve.
[b]PPV: positive predictive value.

### Clinical Implications

Health professionals (ie, physicians, nurses, ICU specialists) can benefit from the advanced accurate predictive capabilities of the intelligent ICU patient monitoring module to help make better decisions regarding major challenges in health care, including bed management, patient flow, stock management, and effective provision of medical supplies. Poor bed management may result in the rejection of new patients, and a reduction in hospital revenue and overall quality of health services [22]. Patient flow involves making decisions regarding admissions, transfers, and referrals. Hospital administration needs solutions that enable reducing waste and wait times, and to increase service efficiency and productivity. Such tools need to consider the uncertainty of patients' recovery status. Poor stock management results in resource shortage or expiration, especially in the ICU where care should be delivered promptly. Thus, integrating the predictive functionalities of the intelligent ICU patient monitoring module within existing decision support platforms and clinical workflows may have several practical implications for improving the quality of care and reducing costs.

### Conclusions

In this article, we proposed a novel approach for predictive modeling with reasonable performance based on a combination of ML algorithms and the quantiles approach that utilizes only vital signs available in the patient's profile without having to use external features. Using this quantiles approach, we engineered additional features by calculating the modified means, SDs, and quantile percentages from the baseline vital sign measures, which provided us with a richer dataset to achieve better predictive power in our models. We applied our approach to build two prediction models: one for mortality prediction and another for ICU LOS. Although the accuracy of the mortality model showed minimal improvement, we achieved better results in the LOS model by around 2.7%.

Intelligent ICU patient monitoring is a promising solution that will improve clinical workflows and enable hospitals to deliver higher-quality, cost-effective patient care, and to improve the overall quality of medical services in the ICU. The solution will support ICUs to put steps ahead and "nudge" health care providers to prepare for unexpected general health conditions of patients and better manage ICU facilities [23]. By relying on a minimal set of features that can be continuously collected from both inside and outside hospital systems and without requiring sophisticated medical devices, our predictive models can be used in cloud-based IRPM systems (see Exhibit X [24], a short video demonstrating the tool in action).

Relying on fewer features will be more feasible for realizing ML algorithms in real-world settings. Future directions of this research will involve adding more predictive modeling capabilities to the intelligent ICU patient monitoring module, including ICU readmission, severity level, and next-day patient vital sign measurements. We are currently working on applying this approach to a wider range of hospital systems within different geographic locations. Integrating intelligent ICU patient monitoring within existing clinical workflows and decision support platforms can support many hospitals in improving the quality of care and reducing costs.

## Acknowledgments

This research was partially supported by King Fahad Medical City, Saudi Arabia. The authors would like to thank Drs. Robert L Davis, and Fatma Gunturkun from the Center for Biomedical Informatics, Department of Pediatrics, University of Tennessee Health Science Center, Memphis, Tennessee, for their feedbacks and insights.

## Conflicts of Interest

None declared.

## Abbreviations

**AUROC:** area under the receiver operating characteristic curve
**ICU:** intensive care unit
**IRPM:** Intelligent Remote Patient Monitoring
**kNN:** K-nearest neighbor
**LOS:** length of stay
**LR:** logistic regression
**LSTM:** long short-term memory
**MAE:** mean absolute error
**MIMIC:** Medical Information Mart for Intensive Care
**ML:** machine learning
**MLR:** multiple linear regression
**PPF:** percent point function
**RF:** random forest
**ROC:** receiver operating characteristic
**SAPS:** Simplified Acute Physiology Score
**SpO$_2$:** oxygen saturation
**SVM:** support vector machine
**SVR:** support vector regression
**XGB:** extreme gradient boosting